\documentclass{article}
\PassOptionsToPackage{table,dvipsnames,svgnames,x11names}{xcolor}
\PassOptionsToPackage{numbers}{natbib}

\usepackage[preprint]{neurips_2025}

\usepackage[utf8]{inputenc} 
\usepackage[T1]{fontenc}    
\usepackage{hyperref}       
\usepackage{url}            
\usepackage{booktabs}       
\usepackage{amsfonts}       
\usepackage{nicefrac}       
\usepackage{microtype}      
\usepackage{xcolor}         

\definecolor{url}{RGB}{0,73,147}
\definecolor{mypink}{HTML}{bc4749}
\definecolor{mygray}{HTML}{6c757d}  

\usepackage{bm} 
\usepackage{xspace}
\usepackage{graphicx}
\usepackage{amsmath} 

\newlength\savewidth\newcommand\shline{\noalign{\global\savewidth\arrayrulewidth
  \global\arrayrulewidth 1pt}\hline\noalign{\global\arrayrulewidth\savewidth}}
  \definecolor{Gray}{gray}{0.92}
\definecolor{DarkGray}{gray}{0.5}
\newcolumntype{x}{>{\columncolor{Gray}}c}
\newcolumntype{H}{>{\setbox0=\hbox\bgroup}c<{\egroup}@{}}
\definecolor{LightCyan}{rgb}{0.88,1,1}
\definecolor{altRowColor}{gray}{0.92}
\definecolor{highlightRowColor}{rgb}{0.9, 0.9, 1}

\newcommand{\grayrow}{\rowcolor{Gray}}

\makeatletter
\DeclareRobustCommand\onedot{\futurelet\@let@token\@onedot}
\def\@onedot{\ifx\@let@token.\else.\null\fi\xspace}
\def\eg{\emph{e.g}\onedot} 
\def\ie{\emph{i.e}\onedot}

\makeatother

\usepackage{caption}
\usepackage{enumitem}
\usepackage{multirow}
\makeatletter
\newcommand{\thickhline}{%
    \noalign {\ifnum 0=`}\fi \hrule height 1pt
    \futurelet \reserved@a \@xhline
}
\makeatother

\usepackage{tabulary}
\newcolumntype{I}{!{\vrule width 1pt}}
\newcolumntype{x}[1]{>{\centering\arraybackslash}p{#1pt}}
\newcolumntype{y}[1]{>{\raggedright\arraybackslash}p{#1pt}}
\newcolumntype{z}[1]{>{\raggedleft\arraybackslash}p{#1pt}}

\usepackage{wrapfig}
\usepackage{subcaption}

\usepackage{algorithm}
\usepackage{algpseudocode}  
\usepackage{listings}

\usepackage{makecell}

\usepackage{hyperref}
\definecolor{citecolor}{RGB}{0,113,187}
\hypersetup{
  colorlinks = true,
  citecolor = citecolor,  
  linkcolor = mypink,      
  urlcolor = url,       
  filecolor = black,      
  pdfborder = {0 0 0}     
}

\title{\texorpdfstring{\raisebox{-5pt}{\includegraphics[height=20pt,width=20pt]{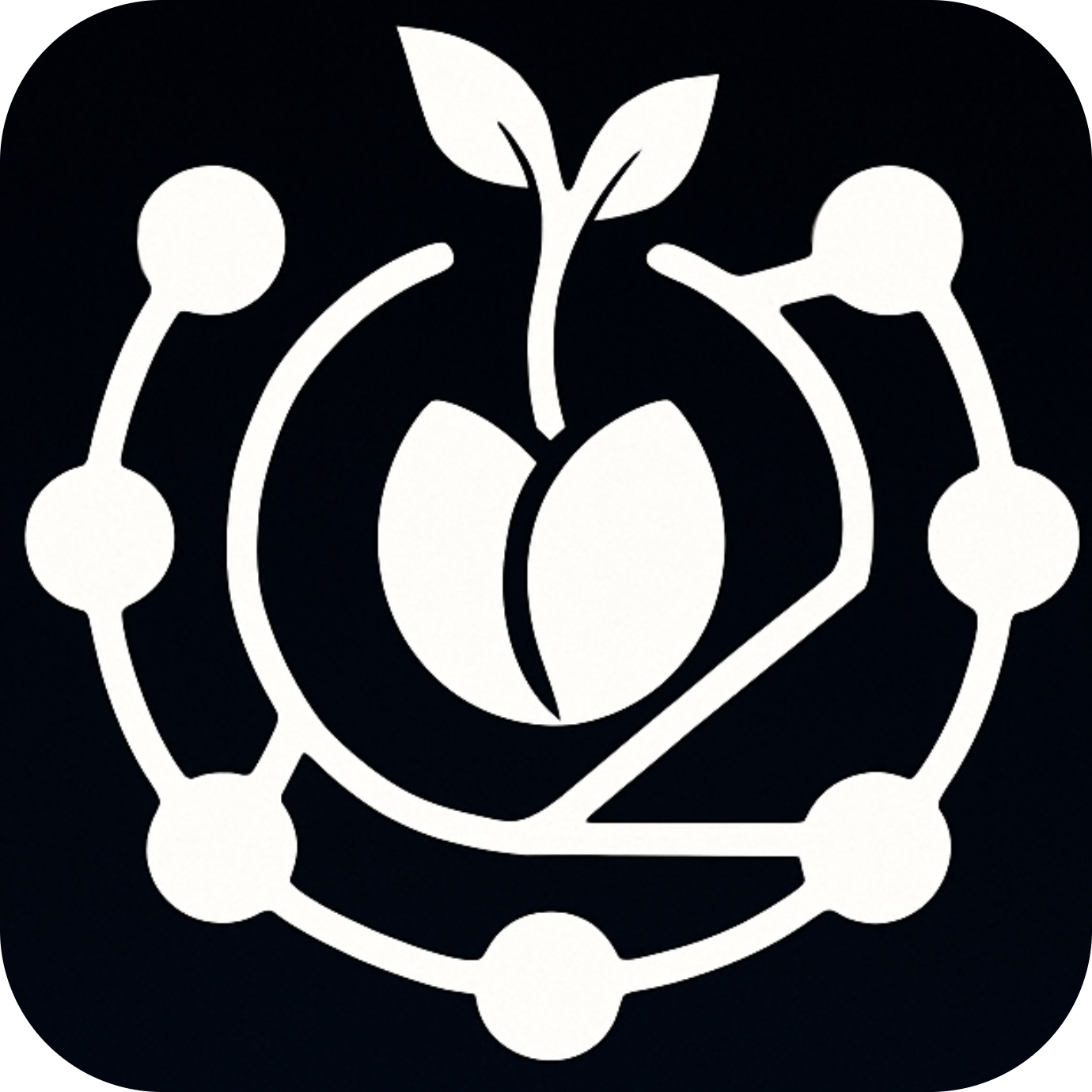}}}{Icon}SEED-GRPO: Semantic Entropy Enhanced GRPO for Uncertainty-Aware Policy Optimization}

\author{Minghan Chen \quad Guikun Chen \quad Wenguan Wang \quad Yi Yang\\
ReLER Lab, CCAI, Zhejiang University \\
}

\begin{document}

\maketitle

\begin{abstract}
Large language models (LLMs) exhibit varying levels of confidence across input prompts (questions): some lead to consistent, semantically similar answers, while others yield diverse or contradictory outputs. This variation reflects LLM's uncertainty about the input prompt, a signal of how confidently the model understands a given problem. However, vanilla Group Relative Policy Optimization (GRPO) \textbf{treats all prompts equally} during policy updates, ignoring this important information about the model's knowledge boundaries. To address this limitation, we propose SEED-GRPO (\textbf{S}emantic \textbf{E}ntropy \textbf{E}nhance\textbf{D} GRPO), which explicitly measures LLMs' uncertainty of the input prompts semantic entropy. Semantic entropy measures the diversity of meaning in multiple generated answers given a prompt and uses this to modulate the magnitude of policy updates. This uncertainty-aware training mechanism enables dynamic adjustment of policy update magnitudes based on question uncertainty. It allows more conservative updates on high-uncertainty questions while maintaining the original learning signal on confident ones. Experimental results on five mathematical reasoning benchmarks (AIME24 \textbf{56.7}, AMC \textbf{68.7}, MATH \textbf{83.4}, Minerva \textbf{34.2}, and OlympiadBench \textbf{48.0}) demonstrate that SEED-GRPO achieves new state-of-the-art performance in average accuracy, validating the effectiveness of uncertainty-aware policy optimization. 
\end{abstract}
  

\section{Introduction}
\label{intro}


Reinforcement learning (RL) emerges as a critical tool for fine-tuning large language models (LLMs)~\cite{shao2024deepseekmath, guo2025deepseek, liu2025understanding, achiam2023gpt, team2025kimi, zhang2025right, lin2025cppo, xiong2025minimalist, zhang2025right, yuan2025vapo, hu2025open,xu2025not} to improve reasoning and accuracy on complex tasks. Leading systems such as OpenAI's GPT-4o and o1~\cite{openai2024o1}, Google's Gemini ~\cite{team2023gemini}, Anthropic's Claude 3 Opus~\cite{anthropic2024claude}, and DeepSeek~\cite{deepseek2024v2,shao2024deepseekmath,guo2025deepseek} all rely on RL techniques to enhance their capabilities beyond what is possible with supervised learning alone. These models demonstrate remarkable proficiency in domains requiring sophisticated reasoning, with RL serving as the key mechanism.

Recent advances like Group Relative Policy Optimization (GRPO)~\cite{shao2024deepseekmath} leverage multiple sampled answers per input prompt to compute relative rewards and advantages within each group, leading to significant gains in reasoning performance. GRPO eliminates the need for a critic model by using the average reward of a group as a baseline, and achieves strong performance on complex reasoning tasks like math and code generation.
\begin{figure}[t]
\centering
\includegraphics[width=0.9\linewidth]{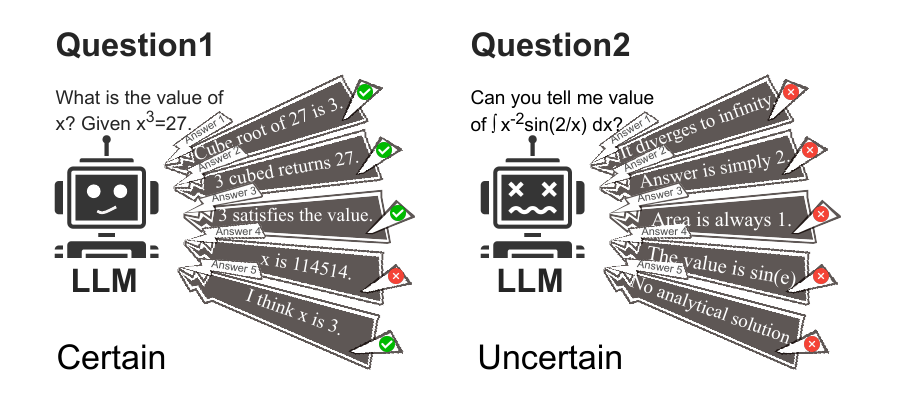}
\caption{Intuitive explanation of semantic entropy. For Question 1, although the 6 responses have slight syntactic variations, 5 of them convey the same meaning, indicating low semantic entropy and high model certainty. For Question 2, the 6 responses can be clustered into 6 distinct meaning classes, resulting in high semantic entropy and indicating significant model uncertainty.}
\label{fig:entropy_example}
\end{figure}

Despite recent progress, GRPO~\cite{shao2024deepseekmath, guo2025deepseek} and its variants~\cite{liu2025understanding, lin2025cppo, yu2025dapo, chu2025gpg, zhang2025grpo, zhang2025srpo} \textbf{assign equal importance to all training prompts} during optimization, ignoring the varying levels of confidence that LLMs demonstrate across different input prompts. However, this signal serves as a crucial probe for model uncertainty, reflecting how well the model understands a given prompt. Uncertainties in responses reveal meaningful information about the model's internal knowledge boundaries. Failing to leverage this uncertainty restricts the ability of policy optimization methods to adaptively focus learning on examples that lie within the model’s current capabilities.

Intuitively, when an LLM model generates highly diverse answers to a given question, it indicates fundamental uncertainty about how to solve the problem. Prior work~\cite{zhang2025right, farquhar2024detecting, wang2023selfconsistency, yao2023tree,   kuhn2023semantic, kossen2024semantic} observes that when an LLM generates semantically diverse answers to the given prompt, it typically indicates the problem lies beyond the model’s current reasoning capability. In other words, such uncertainty may reflect that the question lies outside the model's current reasoning comfort zone, and large policy updates in these cases may result in unstable learning rather than meaningful improvement.  Conversely, when responses demonstrate semantic consistency (\ie, low uncertainty), the model has a coherent understanding of the problem domain, making it safer to update.

This leads us to propose an uncertainty-aware approach to policy optimization: instead of applying uniform updates across all examples, we adaptively adjust the magnitude of updates based on the model's uncertainty for each training prompt. This framework effectively implements a dynamic learning rate mechanism that automatically calibrates according to the model's current capabilities and the problem difficulty. Similar to curriculum learning~\cite{bengio2009curriculum}, this method provides an adaptive learning signal for each problem, reducing the update magnitude for questions that present significant challenges to the model's current reasoning abilities, while maintaining robust training signals for problems where the model demonstrates greater confidence.

In this paper, we introduce SEED-GRPO (\textbf{S}emantic \textbf{E}ntropy \textbf{E}nhance\textbf{D} GRPO), an uncertainty-aware policy optimization algorithm that explicitly quantifies \textbf{prompt-specific uncertainty} using semantic entropy~\cite{kuhn2023semantic, farquhar2024detecting}. Semantic entropy is an entropy-based~\cite{shannon1948mathematical} metric that captures the diversity of meanings among responses to the same input prompt. Figure~\ref{fig:entropy_example}a illustrates this concept. For Question 1, the model produces five correct answers and one incorrect one. While the correct answers vary in syntax and presentation, they all share the same underlying meaning (\ie, $x = 3$), forming a single semantic cluster. This leads to low semantic entropy and indicates low uncertainty. In contrast, for Question 2, the model generates six incorrect and semantically diverse responses, resulting in high entropy and thus high uncertainty. This suggests the model lacks a coherent reasoning path for that problem, possibly because it falls outside the model’s current capabilities. This observed correlation between semantic entropy and problem difficulty provides a principled foundation for uncertainty-aware learning. By leveraging semantic entropy as a proxy for model uncertainty, SEED-GRPO dynamically calibrates policy updates based on how confidently the model responds to each question.

Our contributions can be summarized as follow: \textbf{i)} We propose using \textit{semantic entropy} to quantify model uncertainty at the prompt level, and empirically observe that low entropy correlates with questions within the model's current capabilities (yielding correct predictions), while high entropy signals problems that likely exceed its abilities (resulting in inconsistent and incorrect outputs). \textbf{ii)} We introduce SEED-GRPO, an uncertainty-aware policy optimization algorithm that dynamically calibrates updates based on measured semantic entropy. \textbf{iii)} We demonstrate that SEED-GRPO achieves strong performance across five mathematical reasoning benchmarks (AIME24, AMC, MATH, Minerva, and OlympiadBench), establishing new state-of-the-art results in average accuracy and supporting the effectiveness of semantic entropy-guided optimization.

\section{Related Work}
\textbf{Reasoning LLMs.} 
Recent efforts enhance the reasoning capabilities of Large Language Models (LLMs) through both innovative prompting techniques and sophisticated fine-tuning strategies. Chain-of-Thought~\cite{wei2022chain} prompting encourages models to generate intermediate reasoning steps, resulting in performance improvements on complex mathematical and logical reasoning tasks.
Subsequent research extends this approach through frameworks such as Tree-of-Thoughts~\cite{yao2023tree} and self-consistency CoT~\cite{wang2023selfconsistency}, which strategically aggregate multiple reasoning paths to enhance reliability and accuracy. LIMO~\cite{ye2025limo} demonstrates exceptional results by employing Supervised Fine-Tuning (SFT) to train specialized reasoning models. Meanwhile, alternative approaches to reinforcement learning beyond GRPO show promising outcomes, as evidenced by the effectiveness of Open-Reasoner-Zero~\cite{hu2025open}, KIMI K1.5~\cite{team2025kimi}, and ReST-MCTS*~\cite{zhang2024restmcts}.

\textbf{Group Relative Policy Optimization and Variants.}
DeepSeek first proposes Group Relative Policy Optimization (GRPO)~\cite{shao2024deepseekmath, guo2025deepseek} to train reasoning LLMs. This RL algorithm is a variant of Proximal Policy Optimization (PPO)~\cite{schulman2017proximal} that eliminates the need for value models, which are difficult to train and consume computational resources. GRPO achieves strong performance across numerous reasoning benchmarks spanning mathematics, coding, and question answering domains.
Open-R1~\cite{openr1} represents Hugging Face's fully open-source implementation of GRPO. SRPO~\cite{zhang2025srpo} introduces history resampling, which preserves valuable problems in storage for reuse during later training stages. DAPO~\cite{yu2025dapo} proposes dynamic sampling that filters completely correct and completely incorrect samples to ensure effective training. Dr.GRPO~\cite{liu2025understanding} identifies length bias issues and presents an improved version to address this challenge. Concurrent work on EMPO~\cite{zhang2025right} similarly incorporates semantic entropy, however, they directly incorporate semantic entropy as an optimization objective, whereas our work leverages semantic entropy to measure uncertainty and integrates this uncertainty into advantage calculations. To the best of our knowledge, we are the first to incorporate uncertainty-aware policy optimization into the GRPO framework.

  

\section{SEED-GRPO: Uncertainty-Aware Policy Optimization}
\label{SEED}
\subsection{Motivation: Uncertainty-Aware Learning}

The fundamental insight behind our approach is that when a model generates divergent responses to the same prompt across multiple attempts, such variation often reflects high uncertainty, suggesting that the task potentially exceeds the model's current capabilities (\S\ref{intro}). SEED-GRPO leverages this insight through a principled mechanism: For questions where the model exhibits high semantic entropy (high uncertainty), we adaptively downscale the advantages during policy updates, resulting in more conservative learning steps. This prevents the model from overfitting to potentially noisy rewards on prompts it cannot yet reliably solve. For questions where the model demonstrates low semantic entropy (high certainty), we maintain the original advantages.

This design echoes the principle of curriculum learning~\cite{bengio2009curriculum}, where learning progresses from easier to harder examples. However, rather than relying on static difficulty heuristics, SEED-GRPO employs semantic entropy as a dynamic, model-specific uncertainty signal to calibrate learning pressure.

\begin{figure}[t]
\centering
\includegraphics[width=1\linewidth]{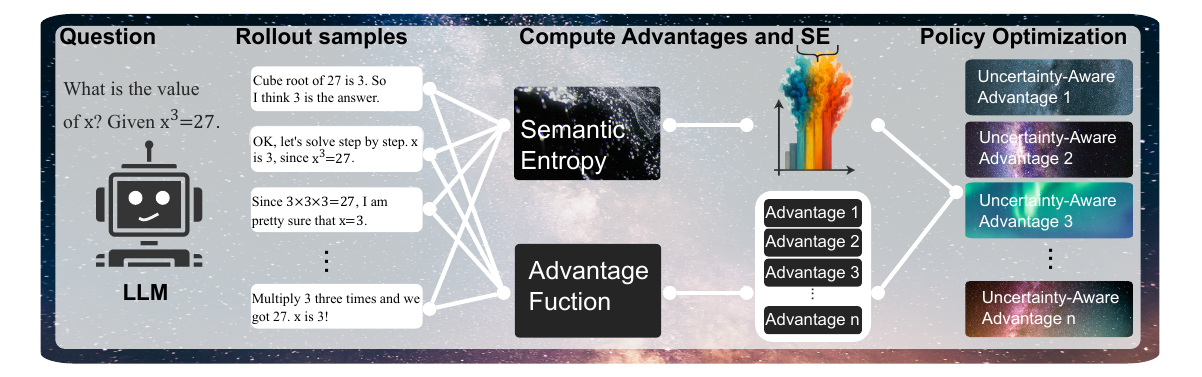}
\caption{The SEED-GRPO framework incorporating semantic entropy for uncertainty-aware reinforcement learning. The framework samples multiple responses from a pre-trained LLM, computes semantic entropy to measure model uncertainty, and modulates the advantage function accordingly to enable more conservative updates for high-uncertainty questions.}
\label{fig:frame}
\vspace{-0.3cm}
\end{figure}

\subsection{SEED-GRPO Illustration via Math Reasoning Example}
To illustrate the core mechanics of SEED-GRPO, consider a math problem $q$ (prompt) such as:
\begin{quote}
\small “What is the value of $x$? Given $x^3=27$.”
\end{quote}
Using an LLM $\pi_{\theta_{\text{old}}}$, we sample a group of $G$ responses $\{\,o_1, o_2, \dots, o_G\,\} \sim \pi_{\theta_{\text{old}}}(\cdot \mid q)$, as shown in  Fig.$_{\!}$~\ref{fig:frame}, responses are also referred as rollout samples. Each response $o_i \in \mathbb{R}^{l_i \times \textit{dim}}$ is a token sequence of length $l_i$ and token dimension $\textit{dim}$. These sequences contain detailed step-by-step reasoning and conclude with a boxed final answer. While such sequences are sometimes referred to as ``trajectories'' in traditional reinforcement learning (\eg, PPO~\cite{schulman2017proximal}), we avoid this terminology for clarity.

Each $o_i$ is an independently sampled text sequence. Some may contain correct solution paths, while others may contain logical or arithmetic errors. We extract the final answers and compute rewards: $r_i = 1$ if $o_i$ is correct, and $r_i = 0$ otherwise. Note that, in  SEED-GRPO there is no reward model, these rewards are obtained by comparing with ground truth labels using specific verification rules. The average group reward $\bar{r} = \frac{1}{G} \sum_{i=1}^G r_i$ serves as a baseline, and advantages can be calculated:
\begin{equation}
\vspace{-0.1cm}
A_i = r_i - \bar{r}, \quad A_i \in \mathbb{R}.
\label{eq:advantage}
\vspace{-0.1cm}
\end{equation}
In SEED-GRPO, the advantage $A_i$ is broadcast across all tokens in the response $o_i$. For instance, if $o_3$ consists of 50 tokens and its advantage is $0.5$, then each of those 50 tokens is associated with the same scalar advantage during training. Once the advantages are computed, policy updates are performed using the clipped surrogate objective inspired by PPO:
\begin{equation}
\vspace{-0.1cm}
\mathcal{L}_i(\theta)=
\min\!\Big(
    \text{ratio}_{i}(\theta)\, A_i,\;
    \text{clip}\bigl(\text{ratio}_{i}(\theta),\,1-\epsilon,\,1+\epsilon\bigr)\, A_i
\Big), \qquad \epsilon=0.2,
\label{eq:loss}
\vspace{-0.1cm}
\end{equation}
where $\text{ratio}_i(\theta) = \frac{\pi_{\theta}(o_i \mid q)}{\pi_{\theta_{\text{old}}}(o_i \mid q)}$ is the importance sampling ratio between the current and old policies.

The overall training objective for question $q$ is the mean over all $G$ samples:
\begin{equation}
\mathcal{L}(\theta) = \frac{1}{G} \sum\nolimits_{i=1}^{G} \mathcal{L}_i(\theta).
\end{equation}
We then update the model parameters by maximizing this objective.

To incorporate uncertainty into the learning process, we measure the \textbf{semantic entropy} $\text{SE}(q)$~\cite{kuhn2023semantic, farquhar2024detecting} of the generated answer group (rollout samples Fig.$_{\!}$~\ref{fig:frame}). Semantic entropy quantifies the degree of semantic diversity across the generated responses. It captures whether the outputs consistently converge on a single reasoning path or instead diverge into multiple, potentially conflicting solutions.

Intuitively, semantic entropy measures how diverse and inconsistent the model's rollout samples are in terms of their meaning. As illustrated in Fig.$_{\!}$~\ref{fig:entropy_example}, for a given mathematical problem, an LLM may generate multiple answers. For Question 1, although there are 6 responses with different syntactic structures and phrasing, 5 of them express essentially the same meaning $x=3$, indicating that the model is highly certain about its answer. Conversely, for Question 2, the 6 responses fall into 6 distinct meaning classes, resulting in high semantic entropy, which suggests that the model struggles with this more challenging problem and lacks confidence in its outputs.

Before computing semantic entropy, we first group the $G$ sampled responses $\{o_1, o_2, \dots, o_G\}$ into a set of meaning clusters $\mathcal{C} = \{C_1, C_2, \dots, C_K\}, \text{where } C_k = \{ o_i : \text{meaning}(o_i) = k \}$:

The semantic entropy is theoretically defined as:
\begin{equation}
\text{SE}(q) = -\sum_c\big((\sum_{o_i \in c} p(o_i \mid q)) \log \big[\sum_{o_i \in c} p(o_i \mid q)\big]\big),
\end{equation}
where $c$ indexes the semantic equivalence classes, and $p(o_i \mid q)$ is the probability of generating response $o_i$ given question $q$ under the current policy model $\pi_{\theta_{\text{old}}}$.

However, in practice, an LLM can generate an unbounded number of diverse responses to a given question $q$, potentially spanning a vast—and unknown—set of meaning classes. It is infeasible to enumerate all possible semantic clusters.

Therefore, we approximate the semantic entropy using the Monte Carlo method proposed in~\cite{farquhar2024detecting,kuhn2023semantic}. Before computing semantic entropy, we first group the $G$ sampled responses $\{o_1, o_2, \dots, o_G\}$ into a set of meaning clusters $\mathcal{C} = \{C_1, C_2, \dots, C_K\}, \text{where } C_k = \{ o_i : \text{meaning}(o_i) = k \}$.
The clustering method will be detailed in \S\ref{exp_setup}. Each cluster $C_k$ represents a semantically coherent subset of responses that share the same meaning, despite possible differences in wording or reasoning steps. 

Based on observed meaning clusters $\mathcal{C} = \{C_1, C_2, \dots, C_K\}$, derived from a finite number of $G$ sampled responses, the semantic entropy of prompt $q$ is estimated as:
\begin{equation}
\text{SE}(q) \approx -\frac{1}{K} \sum\nolimits_{k=1}^{K} \log p(C_k \mid q),
\label{eq:mc-entropy}
\end{equation}
where $p(C_k \mid q) = \sum_{o_i \in C_k} \pi_{\theta_{\text{old}}}(o_i \mid q)$ denotes the total probability mass assigned to the $k$-th observed cluster. This formulation aligns with the Rao–Blackwellized Monte Carlo estimate proposed in~\cite{farquhar2024detecting,kuhn2023semantic}, which approximates semantic entropy over sampled outputs and observed clusters.

Semantic entropy is non-negative and measures model's uncertainty on the given prompt. When all $G$ responses convey the same meaning ($K$=1), the entropy reaches its minimum value of 0, indicating complete certainty. Conversely, when each response belongs to a distinct semantic cluster ($K=G$), the entropy reaches its maximum value, signaling extreme uncertainty where the model produces entirely different answers each time. Given a fixed number of responses $G$, the maximum possible semantic entropy can be calculated as: $\text{SE}_{\max} = \log G.$ For instance, if $G=8$ and all eight responses fall into different semantic clusters, the maximum semantic entropy would be approximately \textbf{2.07}.

This semantic entropy allows us to quantify the uncertainty of the model for each question. Higher entropy indicates greater semantic diversity in the model's responses, suggesting that the model is uncertain about the correct answer. Lower entropy indicates greater consensus among responses, suggesting higher confidence in the model's answers. We leverage this uncertainty measurement to modulate the advantage in the reinforcement learning objective. The key insight is that model updates should be more conservative for questions where the model exhibits high uncertainty. Our uncertainty-aware advantage modulation function is defined as:
\begin{equation}
\hat{A}_i = A_i \cdot f\big(\alpha \cdot\text{SE}(q)/\text{SE}_{\max}(q)\big),
\end{equation}
where $\alpha$ is a hyperparameter controlling sensitivity. When semantic entropy is high, we interpret it as model uncertainty and scale down the advantage to produce more conservative updates. The function $f$ can take various forms, such as linear, exponential, or focal styles, influencing how uncertainty affects the advantage scaling. We conduct ablation studies on $f$ in detail (\S\ref{ablation}).

This modulation attenuates advantages for high-entropy questions, effectively reducing the magnitude of parameter updates for problems where the model is uncertain. Intuitively, this approach makes the training process more cautious about learning from feedback on questions where the model lacks confidence, mitigating the risk of overfitting to potentially noisy or misleading reward signals.

\subsection{Discussion and Analysis}
\textbf{1) Why not use vanilla information entropy to measure uncertainty?}
Shannon entropy~\cite{shannon1948mathematical}, while widely used, can yield misleading estimates of uncertainty in the context of language model outputs. Specifically, when a model generates several responses that differ in phrasing, syntax, or word choice but convey the same underlying meaning, vanilla entropy will still report a high value. This is because it operates on surface-level token distributions and is agnostic to semantic equivalence. Consequently, it overestimates uncertainty in cases where the model is, in fact, semantically consistent. In contrast, semantic entropy clusters responses based on meaning rather than form. This makes semantic entropy a more faithful and robust indicator of a model’s uncertainty on the input prompt.

\textbf{2) What benefits does uncertainty‑aware advantage bring to policy optimization?}

Incorporating uncertainty into the advantage computation allows SEED-GRPO to modulate the learning process adaptively. To better understand this mechanism, we present a simplified gradient analysis of the policy update. For clarity, we consider the loss function without clipping:

\begin{equation}
\mathcal{L}_i(\theta)=\text{ratio}_i(\theta) \cdot \hat{A}_i 
= \frac{\pi_\theta(o_i \mid q)}{\pi_{\theta_{\text{old}}}(o_i \mid q)} \cdot \hat{A}_i.
\end{equation}

The gradient is computed as:
\begin{equation}
\nabla_{\theta} \mathcal{L}_i(\theta)
= \nabla_{\theta} \log \pi_{\theta}(o_i \mid q) \cdot \text{ratio}_i(\theta) \cdot \hat{A}_i.
\end{equation}

Accordingly, the policy update becomes (with the global learning rate $\eta$):
\begin{equation}
\label{eq9}
\theta \leftarrow \theta + \eta \cdot \nabla_\theta \log \pi_\theta(o_i \mid q) \cdot \text{ratio}_i(\theta) \cdot 
\underbrace{\big[ A_i \cdot f\left( \alpha \cdot \text{SE}(q)/\text{SE}_{\max}(q) \right) \big]}_{\hat{A}_i}.
\end{equation}

\vspace{-0.4cm}
By integrating semantic uncertainty into $\hat{A}_i$, this formulation effectively scales the gradient for each input based on the model’s uncertainty. This uncertainty-aware advantage computation effectively creates a question-specific adaptive learning rate. 

As shown in Eq.~\ref{eq9}, the policy update is governed by four components: the global learning rate $\eta$, the log-probability gradient, the importance sampling ratio, and the advantage term. By incorporating the uncertainty-dependent factor $f(\cdot)$, which is non-negative, SEED-GRPO effectively modulates the update magnitude in proportion to the model’s uncertainty. This can be viewed as dynamically adjusting the effective learning rate on a per-question basis.

This mechanism creates an implicit curriculum learning effect: the model naturally takes larger learning steps on problems it can confidently solve, while proceeding more cautiously on challenging ones where the reward signal may be less reliable. This approach helps prevent overfitting to noise in difficult problems while allowing efficient learning from well-understood ones.

\section{Experiments}
\label{exp}
\subsection{Experimental Setup}
\label{exp_setup}

\begin{wraptable}{r}{0.4\textwidth}
\vspace{-0.5cm}
  \caption{Dataset statistics.}
  \label{tab:math-datasets}
  \centering
  \vspace{-0.25cm}
  \resizebox{0.4\textwidth}{!}{%
  \renewcommand\arraystretch{1}
  \begin{tabular}{lcc}
    \thickhline
    \rowcolor[HTML]{f8f9fa}Dataset & \#Questions & Level \\
    \hline
    \hline
    \rowcolor[HTML]{f8f9fa}
    \multicolumn{3}{l}{\textit{Train Datasets}} \\
    MATH (L3–L5) & 8.5k & – \\
    \rowcolor[HTML]{f8f9fa}
    \multicolumn{3}{l}{\textit{Test Datasets}} \\
    AIME24 & 30 & Olympiad \\
    AMC & 83 & Intermediate \\
    MATH500 & 500 & Advanced \\
    Minerva & 272 & Graduate \\
    OlympiadBench & 675 & Olympiad \\
    \bottomrule
  \end{tabular}}
  \vspace{-0.35cm}
\end{wraptable} 
\textbf{Train Datasets.}
 Our training dataset is MATH~\cite{hendrycks2021measuring} Level 3-Level 5, following the same setting of Dr.GRPO~\cite{liu2025understanding}.

\textbf{Test Datasets.}
We evaluate our method on five mathematical reasoning benchmarks: \textbf{i)} AIME24 contains 30 high-school level olympiad problems from the American Invitational Mathematics Examination 2024; \textbf{ii)} AMC includes 83 problems from the AMC series, consisting mostly of multiple-choice questions of intermediate difficulty; \textbf{iii)} MATH500 is a randomly selected subset of 500 problems from the original MATH~\cite{hendrycks2021measuring} dataset, covering algebra, geometry, and number theory; \textbf{iv)} Minerva (MIN)~\cite{lewkowycz2022solving} comprises 272 questions introduced by the Minerva benchmark mostly requiring multi-step reasoning; \textbf{v)} OlympiadBench (OLY)~\cite{huang2024olympicarena} includes 675 high-difficulty math problems.

\textbf{Model.} Following previous works~\cite{guo2025deepseek, liu2025understanding}, we use Qwen2.5-Math~\cite{yang2024qwen2} 1.5B, 7B, and DeepSeek-R1-Distill-Qwen-7B~\cite{guo2025deepseek} as our base models. We choose Dr.GRPO~\cite{liu2025understanding} as the default baseline algorithm.

\textbf{Competitor.} We compare against state-of-the-art methods including Dr.GRPO~\cite{liu2025understanding}, DeepSeek-R1-Zero-Qwen~\cite{shao2024deepseekmath}, RAFT++~\cite{xiong2025minimalist}, GPG~\cite{chu2025gpg}, DAPO~\cite{yu2025dapo}, SimpleRL-Zoo~\cite{zeng2025simplerl}, SRPO~\cite{zhang2025srpo}, Eurus~\cite{yuan2024advancing}, OpenReasoner-Zero~\cite{hu2025open}, and QwQ-preview~\cite{yang2024qwen2.5}.

\textbf{Evaluation Metrics.}
To maintain consistency with prior research ~\cite{liu2025understanding, zeng2025simplerl}, we primarily employ the Pass@1 metric for comparative analysis~\cite{chen2021evaluating}. The pass@$k$ metric evaluates whether, among $k$ responses to a given problem, at least one solution passes the test criteria. The Pass@1 scenario, where only a single response is generated, presents a more challenging setting.  For the uncertainty function $f(\cdot)$, we default choose Linear function with $\alpha=0.02$, more ablation studies are in \S\ref{ablation}.


\textbf{Implementation Details.}
Our training configuration and hyperparameter settings follow Dr.GRPO ~\cite{liu2025understanding}. Specifically, we limit the maximum output to 3,000 tokens, and when calculating advantages, we do not normalize by the group reward standard deviation. Similarly, during loss computation, we do not divide by generation length. For semantic entropy clustering, we employ a straightforward approach that only considers whether the final answers generated by the model are identical. Additional implementation details are provided in the supplementary materials.


\begin{table}
  \caption{Pass@1 performance comparison across multiple mathematical reasoning benchmarks. Results marked with \textsuperscript{+} are reported as the mean $\pm$ standard deviation across 3 runs under the same default experimental setting (\S\ref{exp_setup}). Our other results report the best performance.}
  \label{tab:math-results1}
  \centering
  \renewcommand\arraystretch{1.}
  \resizebox{\textwidth}{!}{%
  \begin{tabular}{lcccccc}
    \thickhline
    \rowcolor[HTML]{f8f9fa}
    Method & AIME24 & AMC & MATH & MIN. & OLY. & Avg. \\
    \hline
    \hline
    \rowcolor[HTML]{f8f9fa}
    \multicolumn{7}{l}{\textit{Baseline methods}} \\
    Qwen2.5-Math-base \textbf{1.5B} \raisebox{-1pt}
    {\includegraphics[height=8pt,width=8pt]{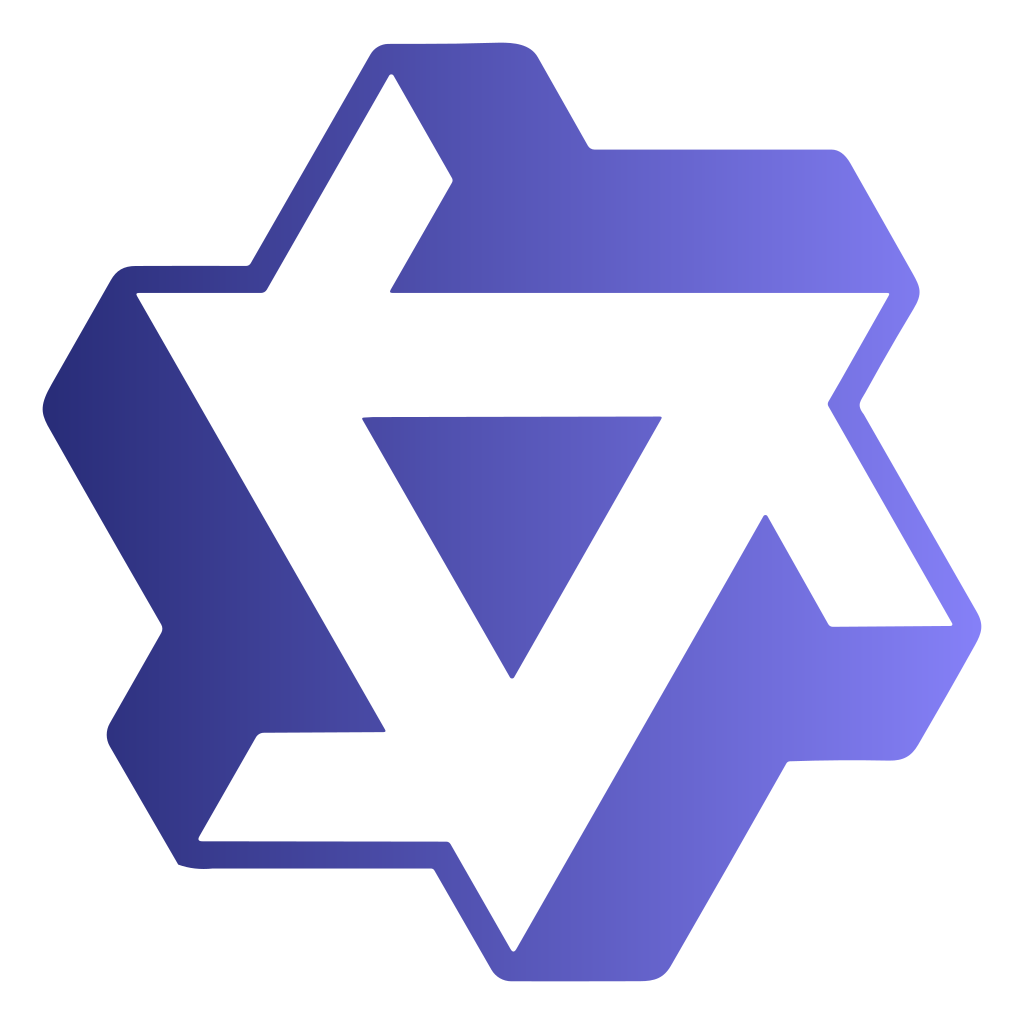}} & 16.7&  43.4& 61.8 &15.1 &28.4 &33.1 \\
    Qwen2.5-Math-base \textbf{7B} \raisebox{-1pt}
    {\includegraphics[height=8pt,width=8pt]{qwen-color.png}} & 0.2 & 45.8 & 69.0 & 21.3 & 34.7 & 38.2 \\
    Dr.GRPO \textbf{1.5B} \raisebox{-1pt}{\includegraphics[height=8pt,width=8pt]{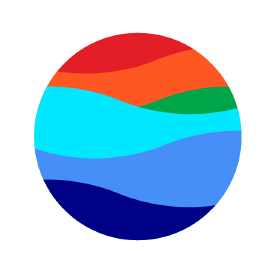}} & 20.0& 53.0& 74.2& 25.7 &37.6 &42.1 \\
    Dr.GRPO \textbf{7B} \raisebox{-1pt}{\includegraphics[height=8pt,width=8pt]{sea.png}} & 43.3 & 62.7 & 80.0 & 30.1 & 41.0 & 51.4 \\
    RAFT++ \textbf{7B} \raisebox{-1pt}{\includegraphics[height=8pt,width=8pt]{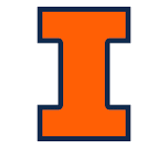}}& - & - & 80.5 & 35.8 & 41.2 & - \\
    OpenReasoner-Zero \textbf{7B} \raisebox{-1pt}{\includegraphics[height=8pt,width=8pt]{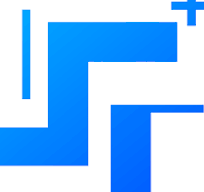}} & 13.3 & 47.0 & 79.2 & 31.6 & 44.0 & 43.0 \\
    Eurus \textbf{7B} \raisebox{-1pt}{\includegraphics[height=8pt,width=8pt]{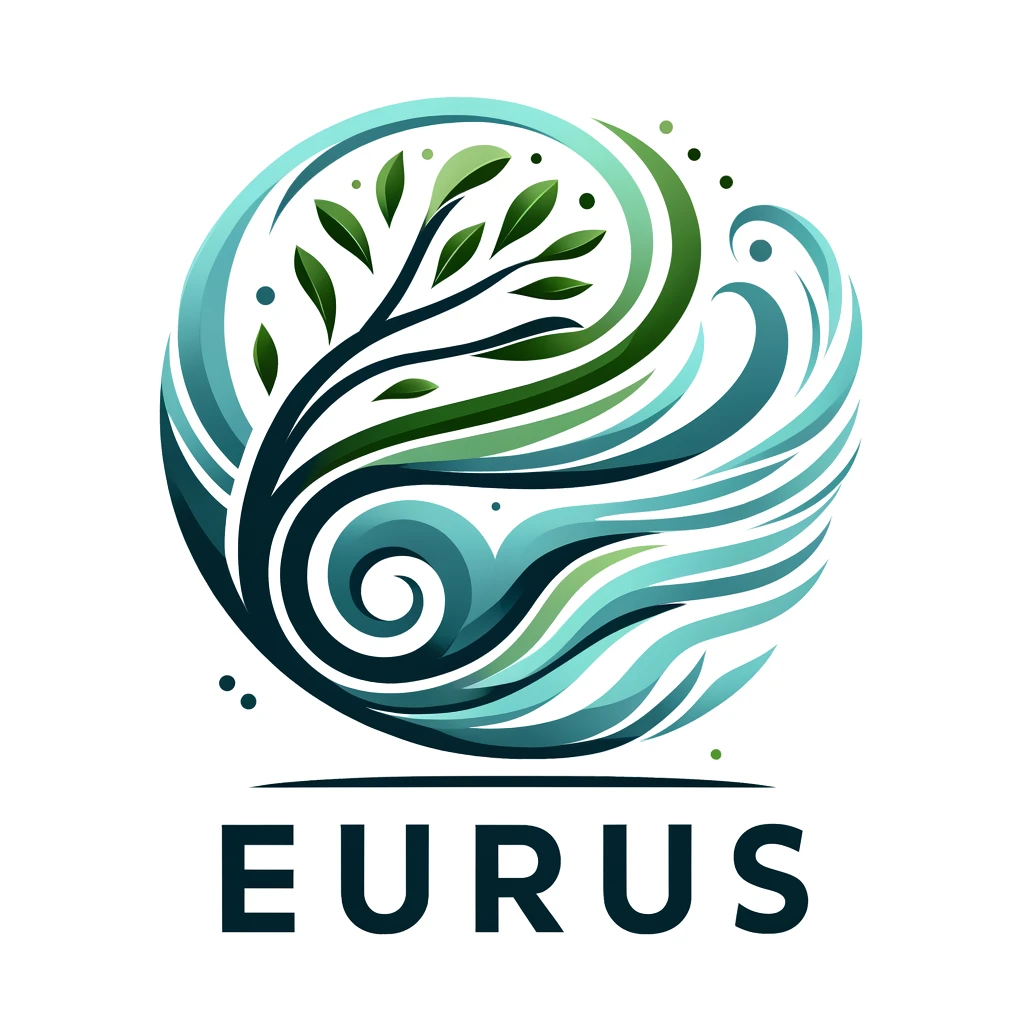}} & 16.7 & 62.7 & 83.8 & 36.0 & 40.9 & 48.0 \\
    SimpleRL-Zoo \textbf{7B} \raisebox{-1pt}{\includegraphics[height=8pt,width=8pt]{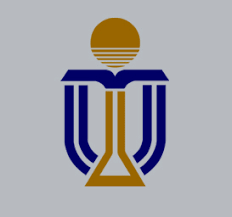}}& 26.7 & 60.2 & 78.2 &27.6 & 40.3 & 46.6 \\
    GPG \textbf{7B} \raisebox{-1pt}{\includegraphics[height=8pt,width=8pt]{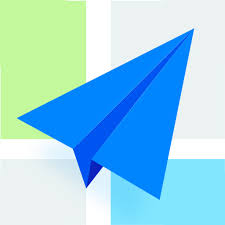}} & 33.3 & 65.0 & 80.0 & 34.2 & 42.4 & 51.0 \\
    SRPO \textbf{32B} \raisebox{-1pt}{\includegraphics[height=8pt,width=8pt]{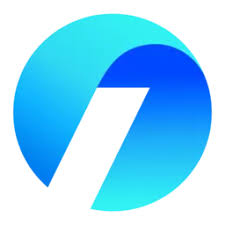}} & 44.3 & - & - & - & - & - \\
    DAPO \textbf{32B} \raisebox{-1pt}{\includegraphics[height=8pt,width=8pt]{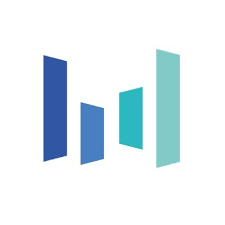}}& \textcolor{mygray}{\small50.0(Avg@32)} & - & - & - & - & - \\
    DeepSeek-R1-Zero-Qwen \textbf{32B} \raisebox{-1pt}{\includegraphics[height=8pt,width=8pt]{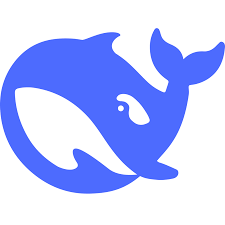}} & 46.7 & - & - & - & - & - \\
    QwQ-preview \textbf{32B} \raisebox{-1pt}{\includegraphics[height=8pt,width=8pt]{qwen-color.png}} & 50.0 & - & 90.6 & - & - & - \\
    \rowcolor[HTML]{f8f9fa}
    \multicolumn{7}{l}{ \textit{Our methods} \raisebox{-1pt}{\includegraphics[height=8pt,width=8pt]{seed-logo.png}}} \\
    SEED-GRPO \textbf{1.5B} (Linear, $\alpha$=0.02) & 23.3 & 50.6 & 75.4 & 26.8 & 41.3 & 43.5  \\
    SEED-GRPO \textbf{7B} (Linear, $\alpha$=0.02)\textsuperscript{+} & $43.3_{\pm 3.4}$ & $64.67_{\pm 4.9}$ & $82.2_{\pm 1.4}$ & $35.03_{\pm 1.6}$ & $45.2_{\pm 2.2}$ & $54.73_{\pm 2.0}$ \\
    SEED-GRPO \textbf{7B} (Linear, $\alpha$=0.02) & 46.7 & 69.9 & 83.0 & 36.7 & 46.8 & 56.6  \\
    SEED-GRPO \textbf{7B} (Linear, $\alpha$=0.02, $G$=16) & \textbf{56.7} & 68.7 & \textbf{83.4} & 34.2 & 48.0 & 58.2 \\
    \grayrow SEED-GRPO \textbf{7B} (Linear, $\alpha$=0.02, R1-Distill) & 50.0 & \textbf{78.3} & \textbf{91.6} & \textbf{38.6} & \textbf{61.5} & \textbf{64.0} \\
    \bottomrule
  \end{tabular}}
  \vspace{-0.7cm}
\end{table}
\begin{table}
  \caption{Training configuration and performance comparison of mathematical reasoning methods.}
  \label{tab:math-results2}
  \centering
  \renewcommand\arraystretch{0.6}
  \resizebox{\textwidth}{!}{%
  \begin{tabular}{lcccccc}
    \thickhline
    \rowcolor[HTML]{f8f9fa}
    Method & \#Train Data & \#Prompt Batch Size & \#Rollouts($G$)& \#Steps & AIME24 & MATH \\
    \hline
    \hline
    \rowcolor[HTML]{f8f9fa}
    \multicolumn{7}{l}{\textit{Baseline methods}} \\
    Dr.GRPO \textbf{7B} \raisebox{-1pt}{\includegraphics[height=8pt,width=8pt]{sea.png}} & 8.5k & 128 & 8 & 400 & 43.3 & 80.0 \\
    SimpleRL-Zoo \textbf{7B} \raisebox{-1pt}{\includegraphics[height=8pt,width=8pt]{hkust.png}}& 7.5k & 1024 & 8 &150 & 26.7 & 78.2 \\
    DAPO \textbf{32B}\raisebox{-1pt}{\includegraphics[height=8pt,width=8pt]{bd.png}} & 17k & 512 & 16 & 5.5k & \textcolor{mygray}{\small50.0(Avg@32)} & - \\
    \rowcolor[HTML]{f8f9fa}
    \multicolumn{7}{l}{ \textit{Our methods} \raisebox{-1pt}{\includegraphics[height=8pt,width=8pt]{seed-logo.png}}} \\
    SEED-GRPO \textbf{7B}  & 8.5k & 128 & 8 & 384 & 40.0 & 81.4 \\
    SEED-GRPO \textbf{7B}  & 8.5k & 128 & 8 &  928  & 46.7 & 83.0 \\
    \grayrow SEED-GRPO \textbf{7B}  & 8.5k & 128 & 16 &  360  & \textbf{56.7} & \textbf{83.4} \\
    \bottomrule
  \end{tabular}}
  \vspace{-0.5cm}
\end{table}

\subsection{Quantitative Comparison Results}
\label{quant}
Table~\ref{tab:math-results1} presents a comprehensive evaluation of our SEED-GRPO approach against established mathematical reasoning methods across multiple benchmarks. Our method demonstrates consistent and substantial improvements over strong baseline systems. Under the Qwen-Math-base setting, SEED-GRPO 1.5B shows significant average improvements compared to the Qwen-Math-base1.5B model, achieving 43.5\% average score across all benchmarks. 

For our default configuration (\S\ref{exp_setup}), SEED-GRPO 7B (Linear, $\alpha$=0.02) achieves an excellent average score of 56.6\% across all benchmarks, representing a significant improvement of \textbf{5.2\%} over the Dr.GRPO 7B baseline. Notably, SEED-GRPO 7B even surpasses SRPO 32B on the challenging AIME24 benchmark (46.7\% vs. 44.3\%), despite having only a fraction of the parameters. This configuration particularly excels on the AMC benchmark with a score of 69.9\%, surpassing all other 7B parameter models with the same initial base architecture.

Our experiments further validate the effectiveness of increasing the number of rollouts $G$ per query. As shown in Table~\ref{tab:math-results1}, simply doubling $G$ from 8 to 16 leads to a \textbf{+1.6}\% gain on average score, and a dramatic \textbf{+10}\% jump on AIME24 (from 46.7\% to 56.7\%). This enhanced configuration achieves an average score of 58.2\% across all benchmarks, outperforming several 32B models including SRPO, DAPO, DeepSeek-R1-Zero-Qwen, and QwQ-preview. Importantly, these results come at a significantly lower computational cost compared to training large 32B models.

Notably, in the DeepSeek-R1-Distill-Qwen-7B setting, our SEED-GRPO (7B, R1-Distill) achieves the best overall performance, with an impressive average score of 64.0\% on Pass@1. It outperforms all 7B and even 32B models across key benchmarks like AIME24, MATH, and OlympiadBench.

Table~\ref{tab:math-results2} compares performance across different training configurations. Compared to baseline methods, our SEED-GRPO achieves superior results with similar or even reduced training data size and computational steps. In particular, with $8.5k$ training data and a batch size of 128, by increasing the number of rollouts to 16, the AIME24 score improved to 56.7\% and the MATH score reached 83.4\%, surpassing all other 7B models. 

It is worth highlighting that our SEED-GRPO 7B (Linear, $\alpha$=0.02) achieves superior performance to several 32B models AIME24, demonstrating the effectiveness of our approach. While DAPO reports a higher Avg@32 score of 50.0\%, our method focuses on the more challenging Pass@1 metric.

\begin{table*}[t]
\centering
\begin{minipage}[t]{0.54\linewidth}
    \centering
    \textbf{(a) \small Method Comparison} \\[2pt]
    \resizebox{1.\linewidth}{!}{
    \begin{tabular}{lcccccc}
    \rowcolor[HTML]{f8f9fa}
        Method & AIME & AMC & MATH & MIN & OLY & Avg.\\
        \shline
        Baseline \textbf{7B} \raisebox{-1pt}{\includegraphics[height=8pt,width=8pt]{qwen-color.png}} & 0.2 & 45.8 & 69.0 & 21.3 & 34.7 & 38.2 \\
        Dr.GRPO \textbf{7B} \raisebox{-1pt}{\includegraphics[height=8pt,width=8pt]{sea.png}} & 43.3 & 62.7 & 80.0 & 30.1 & 41.0 & 51.4 \\
        \grayrow  SEED-GRPO & 46.7 & 69.9 & 83.0 & 36.7 & 46.8 & 56.6   \\
    \end{tabular}
    }
\end{minipage}
\hfill
\begin{minipage}[t]{0.45\linewidth}
    \centering
    \textbf{(b) \small SE Weight $\alpha$} \\[2pt]
    \resizebox{\linewidth}{!}{
    \begin{tabular}{ccccccc}
    \rowcolor[HTML]{f8f9fa}
        $\alpha$ & AIME & AMC & MATH & MIN & OLY & Avg. \\
        \shline
        0.01 &  46.7 & 60.2 & 80.6 & 33.5 & 42.7 & 52.7 \\
        \grayrow 0.02 & 46.7 & \textbf{69.9} & \textbf{83.0} & \textbf{36.7} & \textbf{46.8} & \textbf{56.6} \\
        0.03     & \textbf{50.0} & 61.4 & 83.0 & 34.2 & 44.4 & 54.6 \\
    \end{tabular}
    }
\end{minipage}

\vspace{0.3cm} 

\begin{minipage}[t]{0.54\linewidth}
    \centering
    \textbf{(c) \small Weight Fuction $f(\cdot)$} \\[2pt]
    \resizebox{1.\linewidth}{!}{
    \begin{tabular}{lcccccc}
    \rowcolor[HTML]{f8f9fa}
        Func. &  AIME & AMC & MATH & MIN & OLY & Avg. \\
        \shline
        Focal   & 43.3 & 65.1 & \textbf{84.4} & 35.3 & \textbf{47.6} & 55.1 \\
        Exponential     & 43.3 & 66.3 & 82.0 & 35.7 & 44.3 & 54.3 \\
        \grayrow Linear & 46.7 & \textbf{69.9} & 83.0 & \textbf{36.7} & 46.8 & \textbf{56.6} \\
    \end{tabular}
    }
\end{minipage}
\hfill
\begin{minipage}[t]{0.45\linewidth}
    \centering
    \textbf{(d) \small \#Rollouts ($G$)} \\[2pt]
    \renewcommand\arraystretch{1.05}
    \resizebox{\linewidth}{!}{%
    \begin{tabular}{ccccccc}
    \rowcolor[HTML]{f8f9fa}
        $G$ & AIME & AMC & MATH & MIN & OLY & Avg. \\
        \shline
        8  & 46.7 & \textbf{69.9} & 83.0 & 36.7 & 46.8 & 56.6 \\
        10 & 50.0 & 61.4 & \textbf{84.0} & \textbf{37.5} & \textbf{48.1} & 56.2 \\
        \grayrow 16 & \textbf{56.7} & 68.7 & 83.4 & 34.2 & 48.0 & \textbf{58.2}\\
    \end{tabular}%
    }
\end{minipage}

\vspace{0.3cm} 

\begin{minipage}[t]{0.55\linewidth}
    \centering
    \textbf{(e) \small Base Models} \\[2pt]
    \renewcommand\arraystretch{1}
    \resizebox{1\linewidth}{!}{
    \begin{tabular}{lcccccc}
    \rowcolor[HTML]{f8f9fa}
        Method & AIME & AMC & MATH & MIN & OLY & Avg. \\
        \shline
        \rowcolor[HTML]{f8f9fa}
        \multicolumn{7}{l}{\textit{Qwen2.5 \textbf{1.5B} \raisebox{-1pt}{\includegraphics[height=8pt,width=8pt]{qwen-color.png}} }} \\
        \hspace{2mm} Base         & 16.7&  43.4& 61.8 &15.1 &28.4 &33.1 \\
        \hspace{2mm} Dr.GRPO \textbf{1.5B} \raisebox{-1pt}{\includegraphics[height=8pt,width=8pt]{sea.png}} & 20.0& 53.0& 74.2& 25.7 &37.6 &42.1 \\
        \hspace{2mm} SEED-GRPO     & 23.3 & 50.6 & 75.4 & 26.8 & 41.3 & 43.5 \\
        \rowcolor[HTML]{f8f9fa}
        \multicolumn{7}{l}{\textit{Qwen2.5 \textbf{7B} \raisebox{-1pt}{\includegraphics[height=8pt,width=8pt]{qwen-color.png}}}} \\
        \hspace{2mm} Base        & 0.2 & 45.8 & 69.0 & 21.3 & 34.7 & 38.2 \\
        \hspace{2mm} Dr.GRPO \raisebox{-1pt}{\includegraphics[height=8pt,width=8pt]{sea.png}} & 43.3 & 62.7 & 80.0 & 30.1 & 41.0 & 51.4 \\
        \hspace{2mm} SEED-GRPO & \textbf{56.7} & 68.7 & 83.4 & 34.2 & 48.0 & 58.2 \\ 
        \rowcolor[HTML]{f8f9fa}
        \multicolumn{7}{l}{\textit{R1-Distill \textbf{7B}} \raisebox{-1pt}{\includegraphics[height=8pt,width=8pt]{ds.png}}}\\
        \hspace{2mm} Base        & 10.0 & 26.2 & 80.0 & 30.1 & 41.0 & 51.4 \\
        \hspace{2mm} SEED-GRPO & 50.0 & \textbf{78.3} & \textbf{91.6} & \textbf{38.6} & \textbf{61.5} & \textbf{64.0} \\
    \end{tabular}
    }
\end{minipage}
\caption{SEED-GRPO ablations across five math reasoning benchmarks.}
\label{tab:seedgrpo_ablation}
\end{table*}
\subsection{Ablation Study}
\label{ablation}

\textbf{Method Comparison.} Table~\ref{tab:seedgrpo_ablation}(a) compares SEED-GRPO with the initial base model Qwen2.5-Math-base 7B and the baseline Dr.GRPO 7B. It's important to note that both SEED-GRPO and Dr.GRPO start from the same Qwen2.5-Math-base 7B, using identical hyperparameters. Particularly, SEED-GRPO achieves a remarkable 13.4\% improvement over the baseline on AIME (from 43.3\% to 46.7\%). On average, SEED-GRPO outperforms Dr.GRPO by 5.2\% confirming the effectiveness of uncertainty-aware policy optimization.

\textbf{Semantic Entropy Weight.} We investigate the impact of the semantic entropy weight parameter $\alpha$ in Table~\ref{tab:seedgrpo_ablation}(b), which controls how much influence uncertainty has on the training process. Our results indicate that a medium weight value of $\alpha = 0.02$ yields the best overall performance with an average accuracy of 56.6\%. Interestingly, a higher weight ($\alpha = 0.03$) improves performance on the challenging AIME benchmark but slightly decreases performance on other tasks. This suggests that more difficult tasks may benefit from stronger uncertainty weighting, while simpler tasks require less emphasis on uncertainty. Setting $\alpha$ too low (0.01) consistently underperforms, confirming that some degree of uncertainty modeling is beneficial across all benchmarks.

\textbf{Weight Function.} In Table~\ref{tab:seedgrpo_ablation}(c), we evaluate different functional forms for incorporating semantic entropy into our training objective. We compare linear, exponential, and focal weighting functions. The linear weighting function achieves the best overall performance with an average accuracy of 56.6\%, outperforming both alternatives. While the focal function excels on particular benchmarks like MATH (84.4\%) and OLY (47.6\%), it performs less consistently across all tasks. The exponential function shows competitive but generally lower performance, suggesting that more aggressive uncertainty penalization may not be optimal. These results indicate that a simple linear relationship between semantic entropy and policy updates provides the most robust learning signal.

\textbf{Number of Rollouts.} Table~\ref{tab:seedgrpo_ablation}(d) examines how the number of sampled solutions per query ($G$) affects model performance. Increasing $G$ from 8 to 16 improves the average accuracy from 56.6\% to 58.2\%, with particularly gains on the challenging AIME benchmark (from 46.7\% to 56.7\%). This improvement demonstrates that a larger sample size enables more accurate estimation of semantic entropy. However, the performance with $G=10$ shows mixed results, performing best on some benchmarks (MATH, MIN, OLY) but worse on others (AMC), suggesting task-specific optimal sampling strategies. Overall, our findings support using larger rollout numbers when computational resources permit, with diminishing returns likely beyond $G=16$.

\textbf{Base Models.} Table~\ref{tab:seedgrpo_ablation}(e) shows SEED-GRPO's effectiveness across different base models. When applied to Qwen2.5-1.5B, SEED-GRPO improves average performance by 10.4 percentage points (from 33.1\% to 43.5\%). The improvement is even more substantial for Qwen2.5-7B, with a 20.0 percentage point gain (from 38.2\% to 58.2\%). This demonstrates that SEED-GRPO's benefits scale with model size, suggesting that larger models can better leverage uncertainty information during training. We also evaluated SEED-GRPO on the R1-Distill 7B model, achieving strong performance on AMC (71.2\%) and AIME (46.7\%), though complete results were not available for all benchmarks. Figure~\ref{tab:seedgrpo_ablation}(f) illustrates the performance trend during training, showing that SEED-GRPO consistently outperforms baseline methods throughout the training process, with the performance gap widening as training progresses.


\section{Limitation and Future Work}
\label{limit}
\textbf{Limitation.} Our current implementation of SEED-GRPO focuses solely on utilizing the final answers for semantic clustering in mathematical reasoning tasks, without considering the intermediate reasoning steps. This design choice offers simplicity and proves effective for problems with unique, well-defined answers. However, this approach has several limitations.
First, for open-ended problems without unique answers, our current semantic entropy calculation may not adequately capture the diversity of valid reasoning paths. Second, while focusing on final answers works well for mathematical domains with clear correctness criteria, it may be insufficient for domains requiring nuanced evaluation of the reasoning process itself.

\textbf{Future Work.} SEED-GRPO focuses on mathematical reasoning tasks, where the final answer can be explicitly verified. A promising direction for future work is to extend SEED-GRPO to other domains such as multimodal tasks (image-text VQA, images or videos understanding), code generation, and open-ended textual question answering. These domains often require more nuanced semantic understanding and may benefit even more from uncertainty-aware policy optimization.

Another promising avenue is to refine semantic entropy estimation by incorporating intermediate reasoning steps, rather than relying solely on final answers. This would enable more fine-grained uncertainty modeling along the reasoning trajectory, potentially leading to better training dynamics. Future work could incorporate external models such as commercial LLMs (GPT-4o and o1~\cite{openai2024o1}, Gemini ~\cite{team2023gemini}, Claude 3~\cite{anthropic2024claude}) or open-source models (RoBERTa~\cite{liu2019roberta}, SentenceTransformer~\cite{reimers-2019-sentence-bert, reimers-2020-multilingual-sentence-bert}) as additional meaning clustering models.

Lastly, while SEED-GRPO currently applies semantic entropy during training, it would be valuable to explore test-time compute strategies based on entropy signals. For example, semantic entropy could be used to dynamically adjust generation strategies, such as increasing rollout count or triggering fallback mechanisms when the model is uncertain. Such extensions would allow uncertainty-aware reasoning not only during learning but also at inference time.

\section{Conclusion}
We introduce SEED-GRPO, an uncertainty-aware reinforcement learning framework that enhances Group Relative Policy Optimization (GRPO) by integrating semantic entropy into the training objective. By incorporating a measure of semantic uncertainty into the training objective, our method ensures that policy updates are adaptively scaled according to the model's confidence level for each prompt. This leads to more conservative updates on challenging prompts where the model exhibits high uncertainty, while maintaining effective learning on problems the model can confidently solve. Extensive experiments across five challenging mathematical reasoning benchmarks (AIME24 \textbf{56.7}, AMC \textbf{68.7}, MATH \textbf{83.4}, Minerva \textbf{34.2}, and OlympiadBench \textbf{48.0}) show that SEED-GRPO consistently outperforms strong baselines, including Dr.GRPO~\cite{liu2025understanding} and 32B-scale competitors~\cite{yu2025dapo, hu2025open} achieving new state-of-the-art performance using only a 7B model. Our extensive ablation studies confirm the effectiveness of uncertainty-aware policy optimization.

\bibliographystyle{unsrtnat}
\bibliography{seed}

\end{document}